\documentclass[runningheads]{llncs}

\usepackage{amsfonts}       % blackboard math symbols
\usepackage{nicefrac}       % compact symbols for 1/2, etc.
\usepackage{microtype}      % microtypography
\usepackage{amsmath,amssymb}
\usepackage{enumitem}

\usepackage[misc]{ifsym}

\newcommand{\review}{\mathtt{RE\mbox{-}GrievanceAssist}}
\usepackage{orcidlink}

\begin{document}
\title{$\review$: Enhancing Customer Experience through ML-Powered Complaint Management}
\titlerunning{$\review$}
\toctitle{$\review$: Enhancing Customer Experience through ML-Powered Complaint Management}
\author{Venkatesh C  \and
Harshit Oberoi  \and
Anurag Kumar Pandey  \and
Anil Goyal  \and
Nikhil Sikka}
% \inst{1}
\authorrunning{Venkatesh C et al.}

\institute{Housing.com, India\\
% \url{https://housing.com/}\\
\email{\{firstname.lastname\}@housing.com}}
\maketitle              % typeset the header of the contribution
\begin{abstract}

In recent years, digital platform companies have faced increasing challenges in managing customer complaints, driven by widespread consumer adoption. 
% To address this, automating the process using machine learning technologies offers the most efficient solution, promising streamlined operations and an enhanced customer experience. 
This paper introduces an end-to-end pipeline, named $\review$, designed specifically for real estate customer complaint management.
The pipeline consists of three key components: i) response/no-response ML model using TF-IDF vectorization and XGBoost classifier ; ii) user type classifier using fasttext classifier; iii) issue/sub-issue classifier using TF-IDF vectorization and XGBoost classifier.
Finally, it has been deployed as a batch job in Databricks, resulting in a remarkable 40\% reduction in overall manual effort with monthly cost reduction of Rs 1,50,000  since August 2023. \footnote{Demo Video is available at \url{https://www.youtube.com/watch?v=PM4Q3dNTrr4}}

\end{abstract}

\section{Introduction}
\label{sec:intro}

Over the past few years, online real estate platforms have grown significantly. In recent years, these platforms have started providing multiple allied services to their customers, including property listing, home interiors, rental agreements, and rent payment using credit cards. With the increase in the breadth of services offered, the volume of customer complaints and queries has also increased manifold. On average, these platforms receive more than 1000 complaints daily from different kinds of users pertaining to different sets of services. Generally, a team of manual agents handles these complaints and responds to them (within 2 hours)  through the Freshdesk portal, which aggregates complaints from multiple sources like email, mobile application reviews and social media.

The complaint handling process consists of two steps. Firstly, it involves identifying the type of user (owner, broker, developer, service user etc.) and categorizing the complaint into specific issue and sub-issue classes. There are approximately 35 issues (e.g. payment related, package related, etc) and 10 sub-issues (e.g. payment status, need invoice copy, etc) within each issue category. In the second step, an appropriate response is sent to the user based on the specific sub-issue. The nature of the response is highly dependent on the categorization of the issue and sub-issue. The categorization and response may become subjective based on the agent who handles the specific complaint. Also, the agent spends a sizable amount of their time on tickets which doesn’t require any manual intervention. 

 To overcome these challenges and enhance the customer experience on the platform, we have developed an end-to-end pipeline for real estate complaint management ($\review$). For any incoming ticket on the Freshdesk portal, the proposed solution uses TF-IDF\cite{zhang2011comparative} vectorization and XGBoost classifier for classifying whether manual intervention is required or not. Additionally, the pipeline categorizes the ticket into the type of user using a FastText\cite{joulin2016fasttext} model, the type of issue, and sub-issue using a hierarchical XGBoost classifier. The overall pipeline significantly decreases ticket volume by approximately 40\% which are efficiently closed through our pipeline. Additionally, for tickets requiring a response, our model provides user type, issues, and sub-issues, reducing manual efforts by around 50\% at the ticket level. Finally, we have scheduled the pipeline as a batch job in Databricks.
Furthermore, the implemented system has resulted in a monthly cost reduction of approximately Rs 1,50,000  and an overall $40\%$ reduction in manual efforts since August 2023.

\section{Model  Architecture and Demo}
\subsubsection{Architecture and Experimental Results}
\begin{figure}
    \centering
    \includegraphics[scale=0.4]{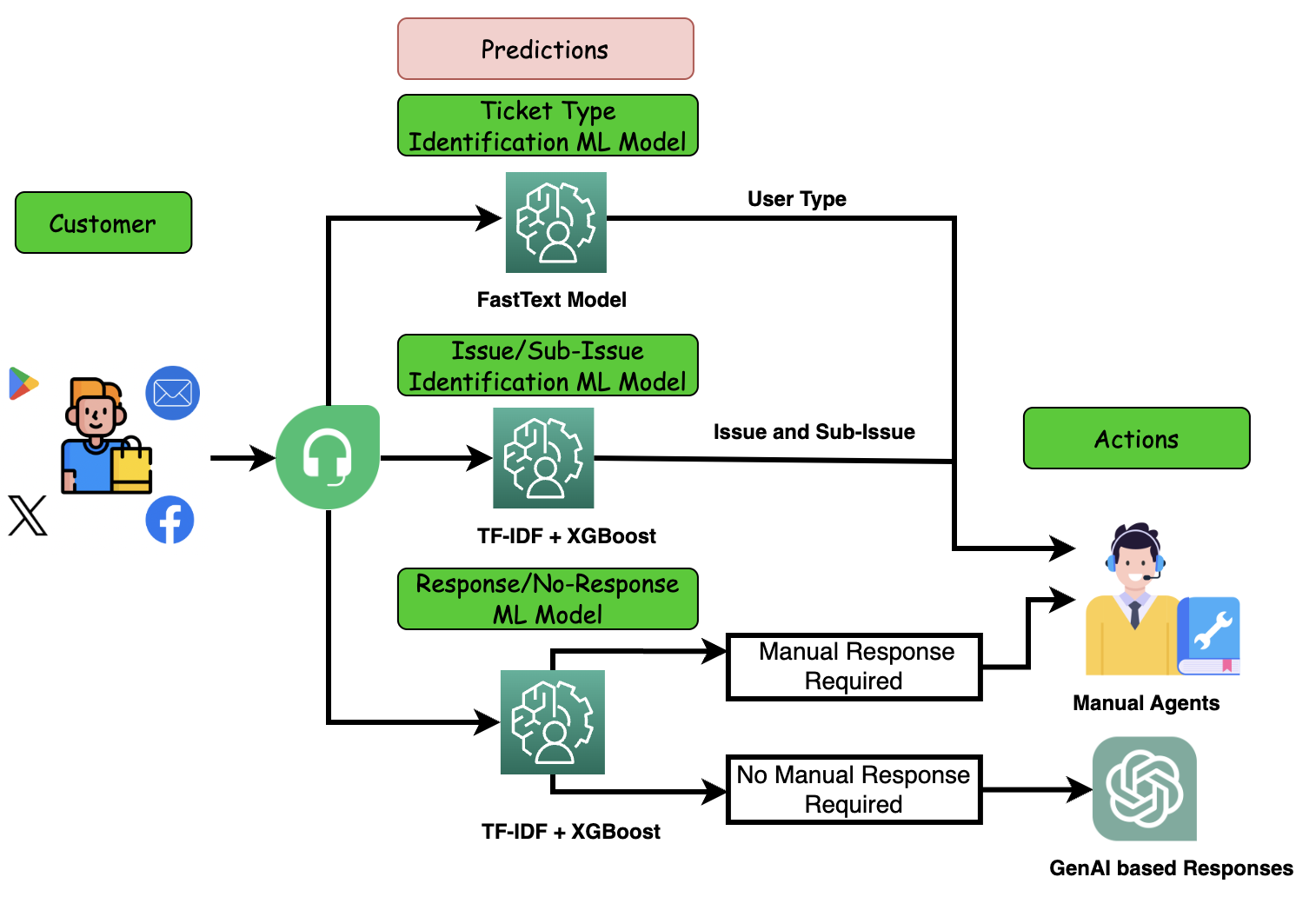}
    \caption{Complete Architecture of $\review$}
    \label{fig:architecture}
\end{figure}

The pipeline has been intricately designed with a modular structure, consisting of three key components: \textit{i)} response/no-response ML model; \textit{ii)} user type classifier; \textit{iii)} issue/sub-issue classifier (as shown in figure \ref{fig:architecture}).
Upon receiving a customer ticket, all three models operate concurrently. We utilize TF-IDF text representation followed by XGBoost classifier \cite{zhang2011comparative} for response/no-response and issue/sub-issue classification. Notably, we adopt a hierarchical approach for issue and sub-issue identification, initially identifying the ticket's main issue followed by sub-issue detection using dedicated models for each issue type. For user type classification, we employ a fasttext classifier\cite{joulin2016fasttext}. Agents respond to tickets manually if required based on information provided by the user type and issue/sub-issue classifiers. When no manual intervention is needed, an automated reply is generated using a predefined template/generative AI-based GPT3 model\cite{brown2020language}.
With this pipeline, we effectively address approximately 40\% of daily tickets without manual intervention. Furthermore, for the remaining 60\%, we reduce manual efforts by 50\% by leveraging insights from the user-type and issue/sub-issue classifiers.

To experimentally evaluate our pipeline, we simulate real-world scenario by using January to June 2023 data for training the models and July 2023 data for testing purposes. 
On the test data, we have obtained F1-score of $86.39\%$, $90\%$, $72.95\%$ and $62.23\%$ for response/no-response ML model, user type classifier, issue classifier and sub-issue classifier respectively.

\subsubsection{Deployment and Demo}

\begin{figure}
    \centering
    \includegraphics[scale=0.34]{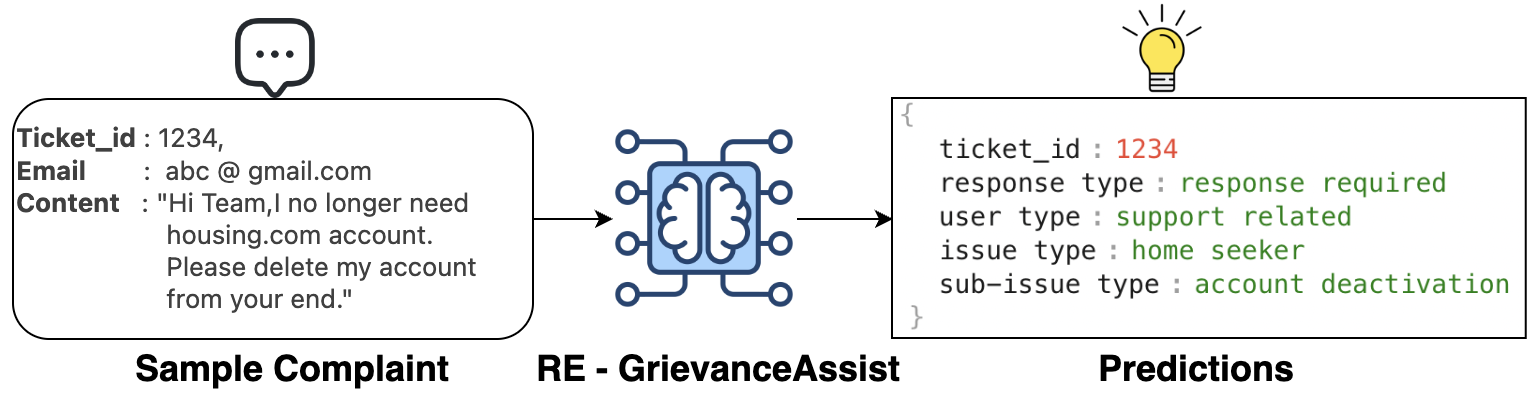}
    \caption{Example Ticket with Response from $\review$}
    \label{fig:sample_ticket}
\end{figure}

The deployment process for our system involves a meticulously designed pipeline that operates as a batch job, scheduled to run at 20-minute intervals. At each iteration, the job retrieves all tickets from the FreshDesk API, initiating the necessary pre-processing steps to ensure data quality and consistency. These pre-processed tickets are then directed to various modules within our system for inference, encompassing response type, user type, issue type, and sub-issue type identification. A sample ticket is shown with its predictions in Figure \ref{fig:sample_ticket}.
\vspace{-5pt}
\vspace{-5pt}
\begin{figure}
    \centering
    \includegraphics[width=1\linewidth]{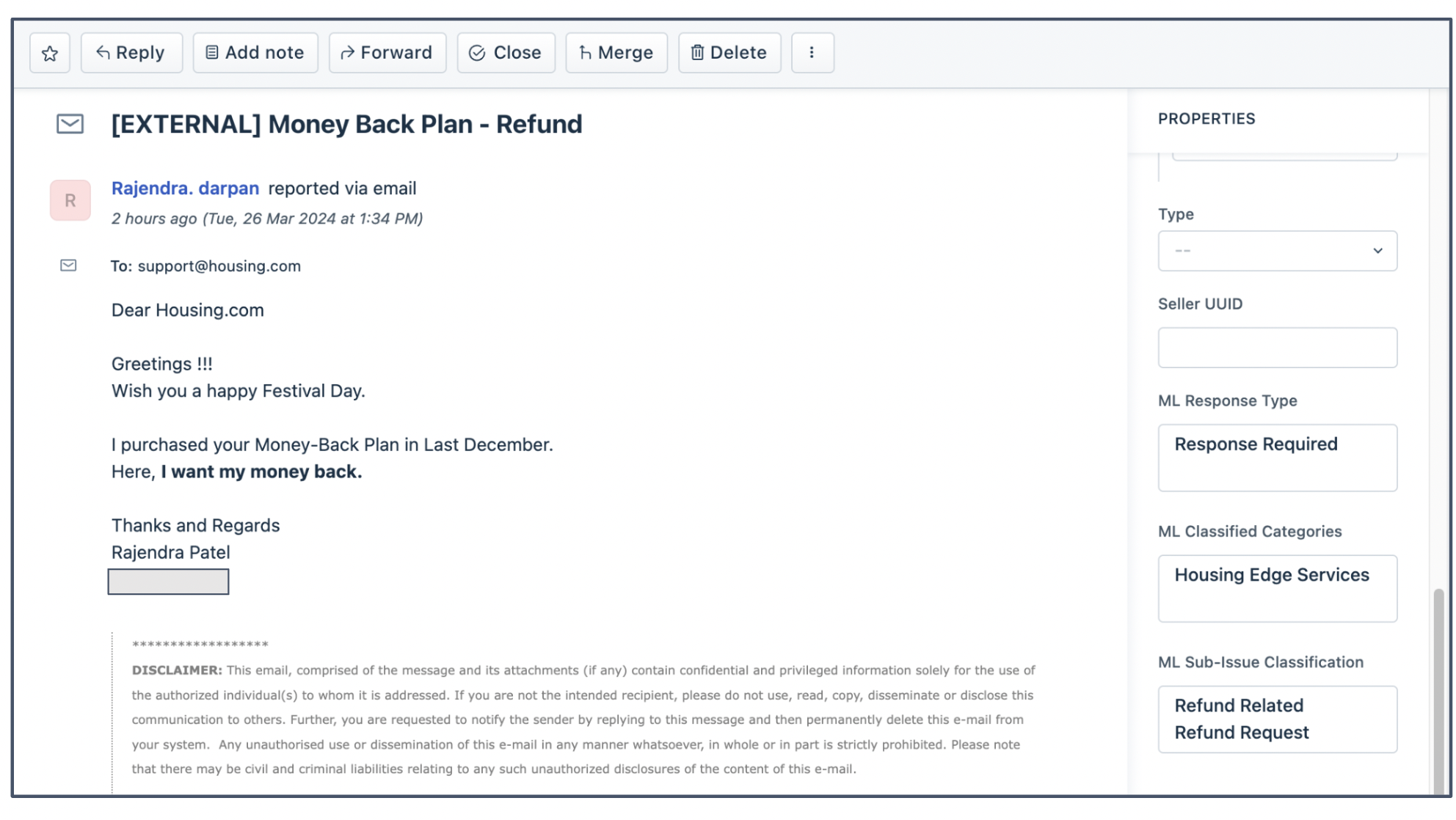}
    \caption{Sample Ticket in Freshdesk UI}
    \label{fig:freshdesk_ui}
\end{figure}
\vspace{-7pt}
The ticket shown in figure \ref{fig:freshdesk_ui} has the customer's complaint on the left and the inferences from our modules are populated on the right-bottom of the image. The response type and user type are populated in ML Response Type and ML Classified Cateogries respectively. Both the issue and sub-issue are populated under the ML-Sub-Issue field in Freshdesk UI. 

Upon completion of inference, standardized template/GPT-3 based responses are automatically dispatched to relevant tickets, streamlining the resolution process. Additionally, all generated results, including inferred classifications and sent responses, are promptly updated both within the FreshDesk API and dedicated log tables, facilitating comprehensive tracking and management of ticket interactions. This systematic deployment approach ensures timely and accurate handling of customer grievances while maintaining seamless integration with our operational workflow. The entire system has been deployed in production since August 2023. \footnote{Demo Video is available at \url{https://www.youtube.com/watch?v=PM4Q3dNTrr4}}

\section{Conclusion}
\label{sec:conclusion}
We present the  $\review$ pipeline for customer complaint management system in real-estate domain. We proposed a machine learning-based automated pipeline tailored for customer complaint management. Our approach robustly identifies response type, user type, issue type, and sub-issue type, including issue-specific responses using a template/GPT3 generative AI model. Through the deployment of a near real-time job, we have  enhanced the speed and accuracy of customer grievance resolution. 
% The deployment in production has led to a significant decrease of 40\% in manual labor and has cut monthly costs by approximately Rs 1,50,000. In future, we plan to explore deep learning methodologies to further enhance the overall performance across all modules.

\bibliographystyle{splncs04}
\bibliography{mybibliography}

\end{document}